\documentclass[runningheads]{llncs}
\usepackage[colorlinks=true, allcolors=blue]{hyperref}

\usepackage{amsmath}
\usepackage[table]{xcolor}
\usepackage{multirow}
\usepackage{booktabs}
\usepackage{enumitem}
\usepackage{graphicx}

\usepackage[skip=2pt]{caption}
\usepackage[colorlinks=true, allcolors=blue]{hyperref}
\usepackage{subcaption}

\usepackage{todonotes}

\begin{document}
\title{Measuring the Quality of Text-to-Video Model Outputs: Metrics and Dataset}
%
%

\author{Iya Chivileva\inst{1} \and Philip Lynch\inst{1} \and Tomás E. Ward\inst{2}\orcidID{0000-0002-6173-6607} \and \\ Alan F. Smeaton\inst{2}\orcidID{0000-0003-1028-8389}}
\authorrunning{Chivileva et al.}
\titlerunning{Text-to-Video Metrics and Dataset}

\institute{School of Computing, Dublin City University, Ireland
\and
Insight Centre for Data Analytics, Dublin City University, Ireland\\
\email{alan.smeaton@dcu.ie}}

\maketitle              
\begin{abstract}

Evaluating the quality of videos generated from text-to-video (T2V) models is important if they are to produce plausible outputs that convince a viewer of their authenticity.  We examine some of the  metrics used in this area and highlight their limitations. 
The paper presents a dataset of more than 1,000 generated videos from 5 very recent T2V models on which some of those commonly used quality metrics are applied. We also include extensive human quality evaluations on those videos, allowing the relative strengths and weaknesses of  metrics, including human assessment, to be compared.
The contribution is an assessment of commonly used quality metrics, and a comparison of their performances and the performance of human evaluations on an open dataset of T2V videos. 
Our conclusion is that naturalness and semantic matching with the text prompt used to generate the T2V output are important but there is no single measure to capture  these subtleties in assessing T2V model output.

\end{abstract}
\section{Introduction}

Recent rapid advances in text-to-video (T2V) generation algorithms have concentrated on improving the state of the art while producing a reliable metric to measure the quality of  generated videos is often an afterthought. It is common for a model to be evaluated on 3 or 4 different metrics and in some cases to also include human assessment. These evaluations show a large variety in the  quality of videos generated from T2V models. Considering this variance, how to fairly evaluate the quality of generated videos is important but the development of quality metrics has not received much attention.

In the field of AI-generated images, recent work by Li {\it et al.}~\cite{li2023agiqa} has compared the quality of images generated from a collection of text-to-image (T2I) models. This used automatically-computed quality metrics and using human evaluations of perception and alignment of the image to the prompt used to generate it. That work also included the release of a database of 2,982 T2I images and the prompts and model parameters used to generate them as well as a comparison between human evaluations and automatically created metrics.
The work in this paper follows a similar sequence to~\cite{li2023agiqa} except that we address evaluating the quality of text-to-video instead of text-to-image models. 

Our aim is assess how the number of metrics needed to produce a reliable evaluation of the output from a T2V model could be reduced. This includes comparing the outputs of automatic metrics with human evaluations. We begin with an examination of open-source state-of-the-art T2V models and   address  the limitations with  current evaluation metrics including  major pitfalls.  We then present the output of human assessment of the authenticity and realism of videos which we refer to as video naturalness, plus the degree to which a generated video aligns with the input prompt which we refer to  as text similarity. We also include the semantic matching between the original text prompt and the content inside the generated video. 
In Section~\ref{sec:Evaluation}, we compare the results of human evaluations with commonly used T2V metrics. Our findings suggest that human evaluations mostly align with  commonly used metrics, but not always so.

\section{Related Work}
\label{sec:Related Works}

\subsection{Text-to-Video Models}
\label{sec:Text-to-Video Models}
In 2022 the first open-source T2V model called Tune-a-Video  was  released by Wu {\it et al.}  \cite{wu2022tune} introducing a mechanism that uses the Stable Diffusion model~\cite{rombach2022high} for video generation. This model is built on state-of-the-art T2I diffusion models and involves a tailored spatio-temporal attention mechanism and an efficient one-shot tuning strategy. 
It served as an inspiration for the rapid development of other open-source models including the following, which are used in this paper.

\begin{enumerate}[nosep]
    \item VideoFusion~\cite{luo2023videofusion} in 2023 uses
a decomposed diffusion process to resolve  per-frame noise as a base noise that is shared among all frames  leading to smoother video output.

\item Text-to-Video Synthesis, also based on the work described in~\cite{luo2023videofusion} in 2023 and openly available 
is also a multi-stage text-to-video generation diffusion model which consists of text feature extraction, a text feature-to-video latent space diffusion model, and video latent space to video visual space. This model adopts the Unet3D structure, and performs video generation through an iterative denoising process from pure Gaussian noise video.

\item Text2Video-Zero~\cite{khachatryan2023text2video} in 2023 takes a low-cost zero-shot text-to-video generation  approach, leveraging the power of  Stable Diffusion and tailoring it for  video. It achieves this by enriching the latent codes of frames with motion dynamics and using a  cross-frame attention of each frame on the first frame to preserve the context, appearance, and identity of  foreground objects.

\item Aphantasia~\cite{aphantasia} also from 2023, is a collection of text-to-image tools, evolved from the artwork of the same name. It is based on the CLIP model and Lucent library, with FFT/DWT/RGB parameterizers and no-GAN generation.

\end{enumerate}

\subsection{Evaluation Metrics}
\label{sec:Evaluation Metrics}
The 3 most commonly used metrics for evaluating video quality are  as follows.

\textbf{1. Inception Score (IS)}~\cite{salimans2016improved} was developed as an  alternative to human evaluation and aims to measure both image quality and diversity. It relies on the ``inception network"~\cite{szegedy2016rethinking} to generate a class probability distribution for images. Higher-quality images should have a low entropy probability \(P(y|x)\), while diversity is measured with the marginal distribution of all  images, which should have high entropy.

\textbf{2. Fréchet Video Distance (FVD)}~\cite{unterthiner2019fvd} measures the distance between  feature activations of real and generated videos in the feature space of a pre-trained video classifier, similar to the approach in the Fréchet Inception Distance (FID)~\cite{heusel2017gans} which was developed for images. A lower FVD score indicates better quality video generation indicating that both the reference and generated videos have similar distributions for feature activations.

\textbf{3. CLIPSim}~\cite{wu2021godiva} uses the CLIP~\cite{radford2021learning} model to evaluate  semantic matching between an initial text prompt and a generated video. CLIP is a contrastive learning-based model that creates a joint embedding space for images and text, allowing the model to understand  relationships between them. CLIPSim extends CLIP to evaluate videos by finding the CLIP score of each frame in a video and returning the average frame score.

\begin{figure}[htb]
    \centering
    \begin{subfigure}{0.49\textwidth}
    \centering
    \includegraphics[width=0.6\textwidth]{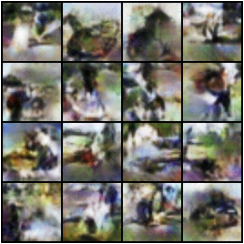}
    \caption{Poor Quality Images from~\cite{barratt2018note}}
    \label{fig:IS_bad}
    \end{subfigure}
    \hfill
    \begin{subfigure}{0.49\textwidth}
    \centering
        \raisebox{10pt}{\includegraphics[width=0.9\textwidth]{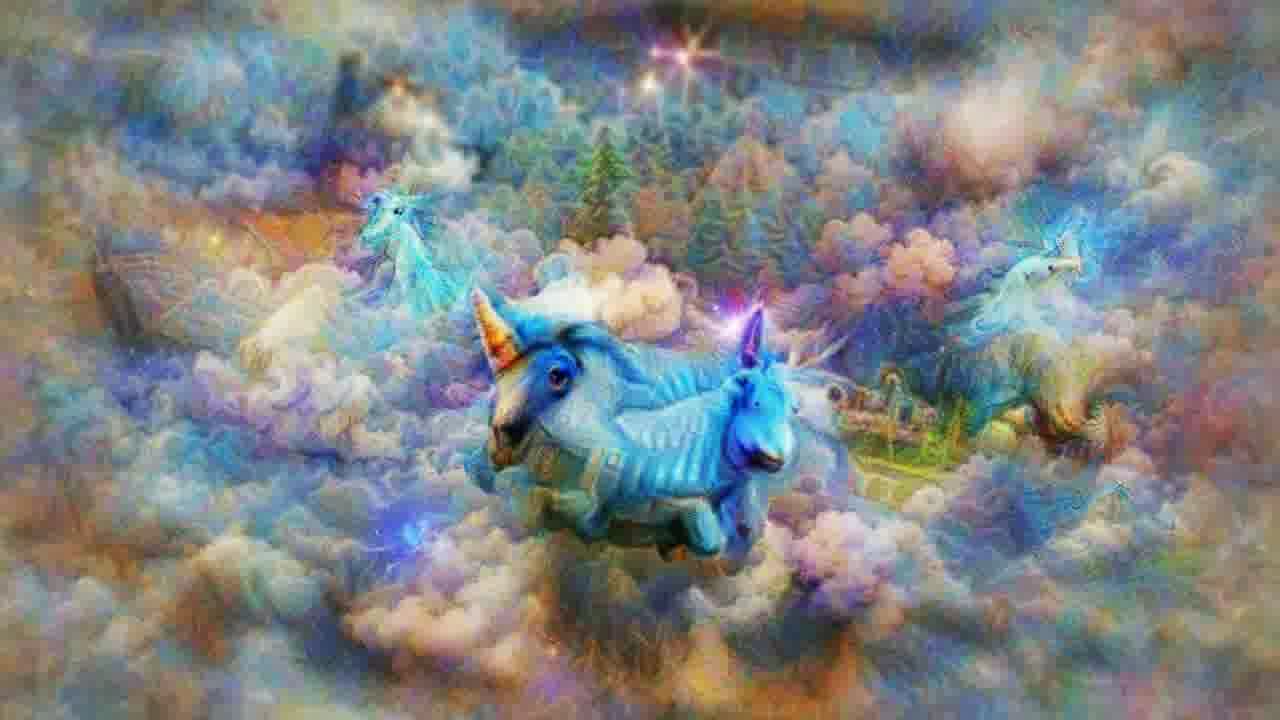}}
        \caption{CLIPSim Error}
        \label{fig:CLIPSim Error}
    \end{subfigure}
\caption{Example limitations of existing T2V quality metrics.}
\label{fig:MetricDrawbacks}
\end{figure}

Although the use of these metrics is common, there are  concerns about their use. \textbf{IS} has been criticised for its tendency to overfit on models trained using ImageNet and  its inability to distinguish between poor and high-quality images~\cite{barratt2018note} as demonstrated in \autoref{fig:IS_bad} where IS assigned an almost perfect score for the examples shown. 
\textbf{FVD} requires reference videos in order to generate a score, which is challenging when comparing T2V models trained on different datasets which hinders the broader applicability of this metric.
The underlying model used in \textbf{CLIPSim}, CLIP, has been criticised by the authors of BLIP~\cite{li2023blip} for its reliance on noisy web image-text pairs arguing that a smaller, filtered dataset should be used. Finally, the ability of Image-to-Text models such as BLIP and CLIP to generate semantically similar captions for images/frames that do not appear to match visually is another concern, as illustrated in \autoref{fig:CLIPSim Error}. Here the video was generated using the Aphantasia T2V model~\cite{aphantasia} with the text prompt ``A blue unicorn flying over a mystical land" and by using CLIPSim we generated a similarity score of over 70\%.

In summary, the metrics are used widely but have limitations which motivates our interest in assessing them and how they compare to human assessments. 

\subsection{Image Naturalness}
\label{sec:Image Naturalness}

Image naturalness refers to how realistic and free of distortions or artefacts an image appears. Naturalness is  related to image quality, which encompasses aspects such as sharpness, contrast, and colour accuracy, but image naturalness specifically focuses on the realism of an image. Non-natural images  lack recognisable and interpretable real-world objects or scenes and  may include computer-generated graphics, abstract art, or heavily manipulated photographs. They  often serve artistic or functional purposes but  do not necessarily reflect the properties of natural images that are easily interpreted by human vision.


The Naturalness Image Quality Evaluator (NIQE) is a no-reference image quality assessment metric~\cite{mittal2012no} based on the observation that natural images tend to exhibit a unit-normal Gaussian characteristic in their luminance values. NIQE uses a set of natural scene statistics (NSS) that captures the statistical regularities present in natural scenes that are not present in unnatural or distorted images.  NIQE  was trained on the LIVE image quality assessment database~\cite{sheikh2006image}, with 29 reference  and 779 distorted images, each with 5 subjective quality scores. 

BRISQUE (Blind/Referenceless Image Spatial QUality Evaluator) is an image quality assessment measure~\cite{mittal2012making} that also uses NSS to evaluate the quality of a distorted image without requiring a reference image. BRISQUE extracts 36-dimensional feature vectors from 96 non-overlapping blocks of the distorted image and maps these onto a reduced-dimensional space using principal component analysis. The quality score is  calculated using a support vector regression (SVR) model trained on the LIVE IQA database~\cite{sheikh2006image}.

\begin{figure}[htb]
    \centering
    \begin{subfigure}[t]{0.19\textwidth}
        \includegraphics[width=\textwidth,height=2cm]{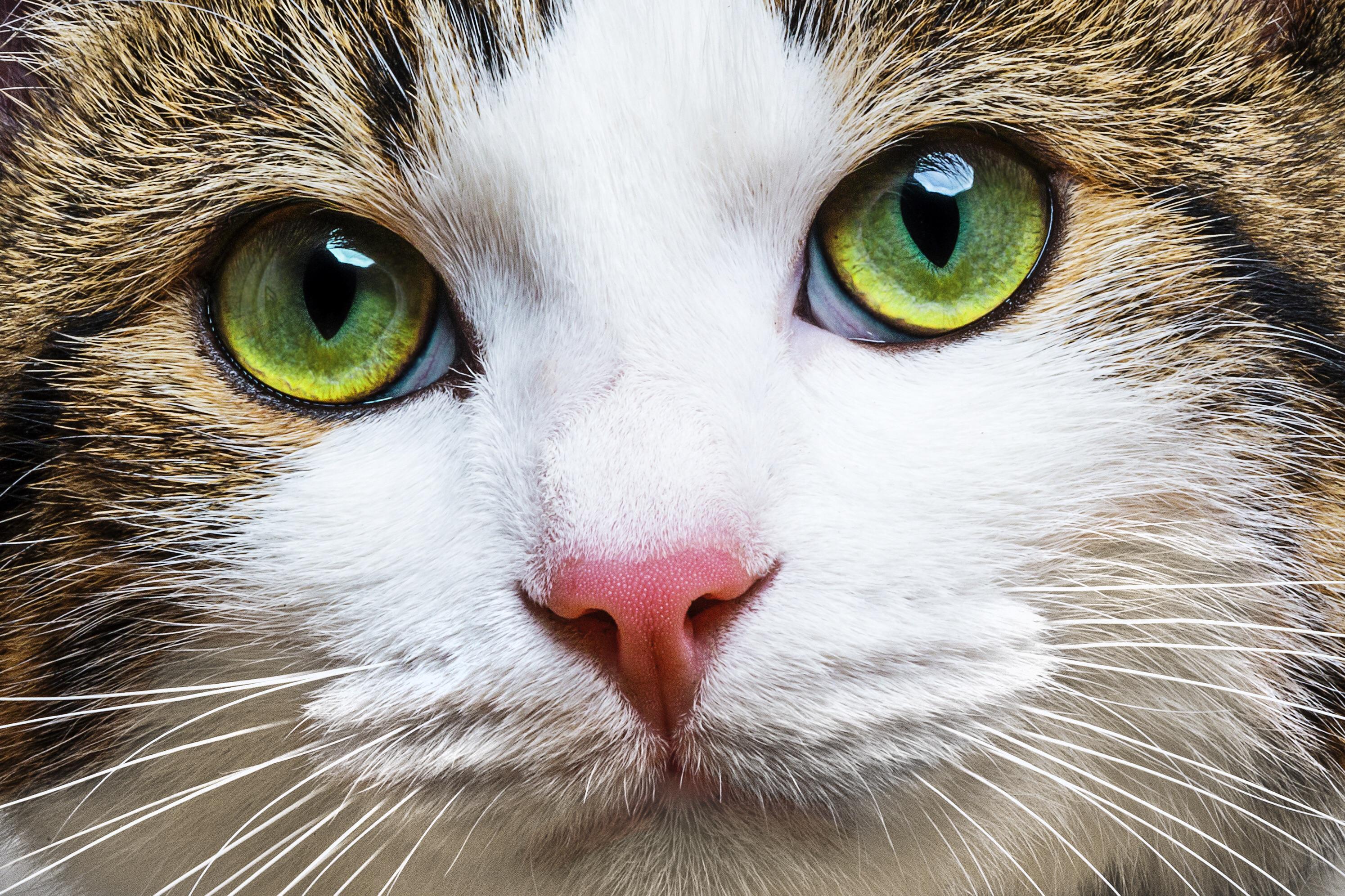}
        \caption{{N:2.5/B:23.8}}
    \end{subfigure}
    \hfill
    \begin{subfigure}[t]{0.19\textwidth}
        \includegraphics[width=\textwidth,height=2cm]{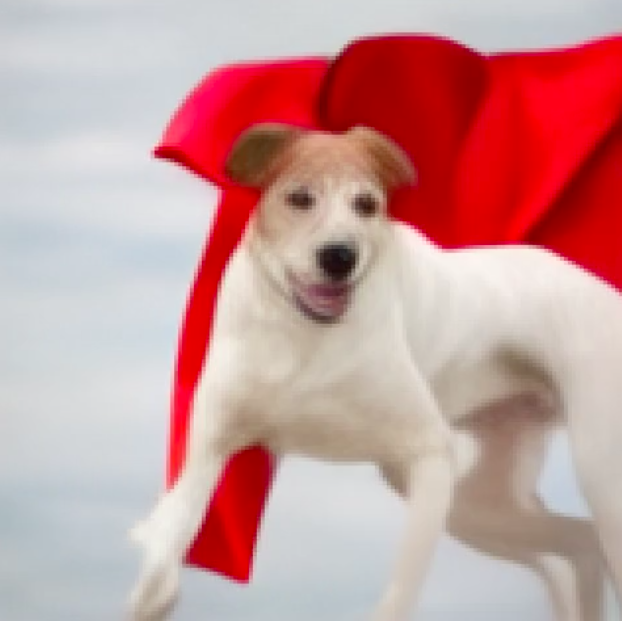}
        \caption{N:7.7/B:54.4}
    \end{subfigure}
    \hfill
    \begin{subfigure}[t]{0.19\textwidth}
        \includegraphics[width=\textwidth,height=2cm]{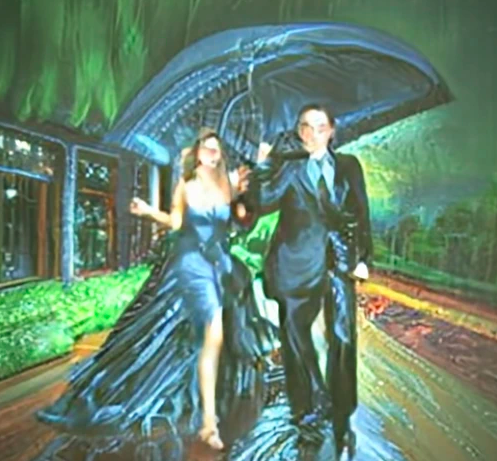}
        \caption{N:5.1/B:39.8}
    \end{subfigure}
    \hfill
    \begin{subfigure}[t]{0.19\textwidth}
        \includegraphics[width=\textwidth,height=2cm]{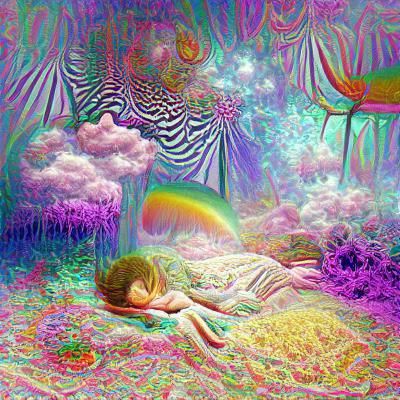}
        \caption{N:3.5/B:15.7}
    \end{subfigure}
    \hfill
    \begin{subfigure}[t]{0.19\textwidth}
        \includegraphics[width=\textwidth,height=2cm]{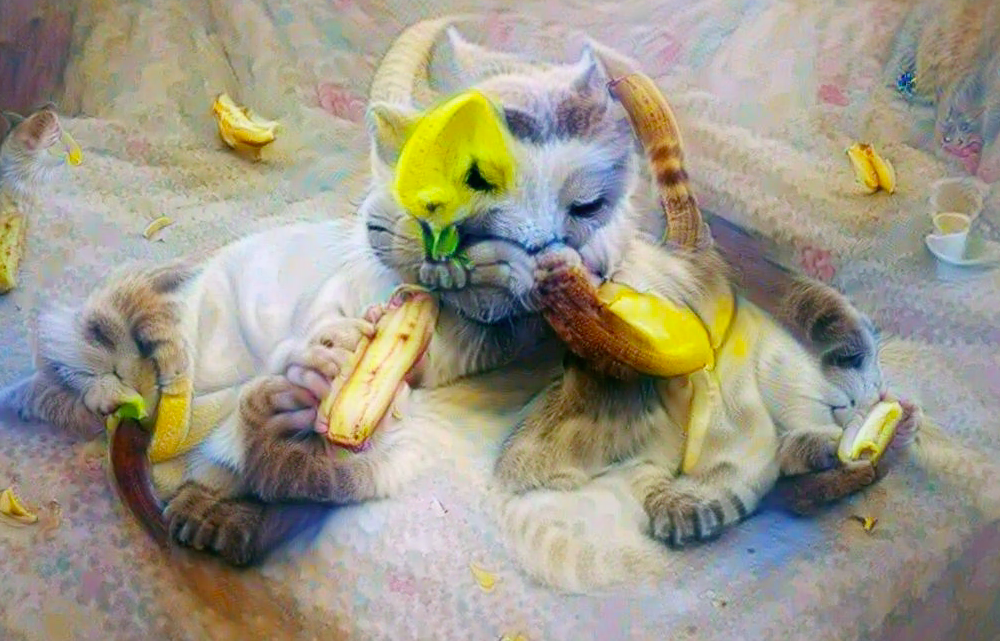}
        \caption{N:3.4/B:22.8}
    \end{subfigure}
    \caption{Image naturalness assessment with NIQE (N) and BRISQUE (B) scores.}
    \label{fig: brisque and niqe results}
\end{figure}

The performance of NIQE and BRISQUE on real photo (a) and frames extracted from various T2V models is shown in  \autoref{fig: brisque and niqe results} where the metrics evaluate images on a scale of 0 to 100,  higher scores indicate lower naturalness. We see  images (b) and (c) received the highest scores indicating poor naturalness and  non-natural images  (d) and (e) received better scores than the low-quality image of a dog in (b) and the image of an oil painting of a couple in (c), which still represent recognisable objects. Although NIQE scores showed slightly better results than BRISQUE, they were still unable to fully differentiate between natural and non-natural images. Based on these properties  we opted to develop a new classifier to detect the naturalness of an image, acknowledging that metrics like NIQE and BRISQUE are primarily concerned with the visual quality of generated videos rather than their naturalness.

\section{An Ensemble Video Quality Metric}
\label{sec:Proposed Metric}

\begin{figure}[htb]
    \centering
    \includegraphics[width=0.75\textwidth]{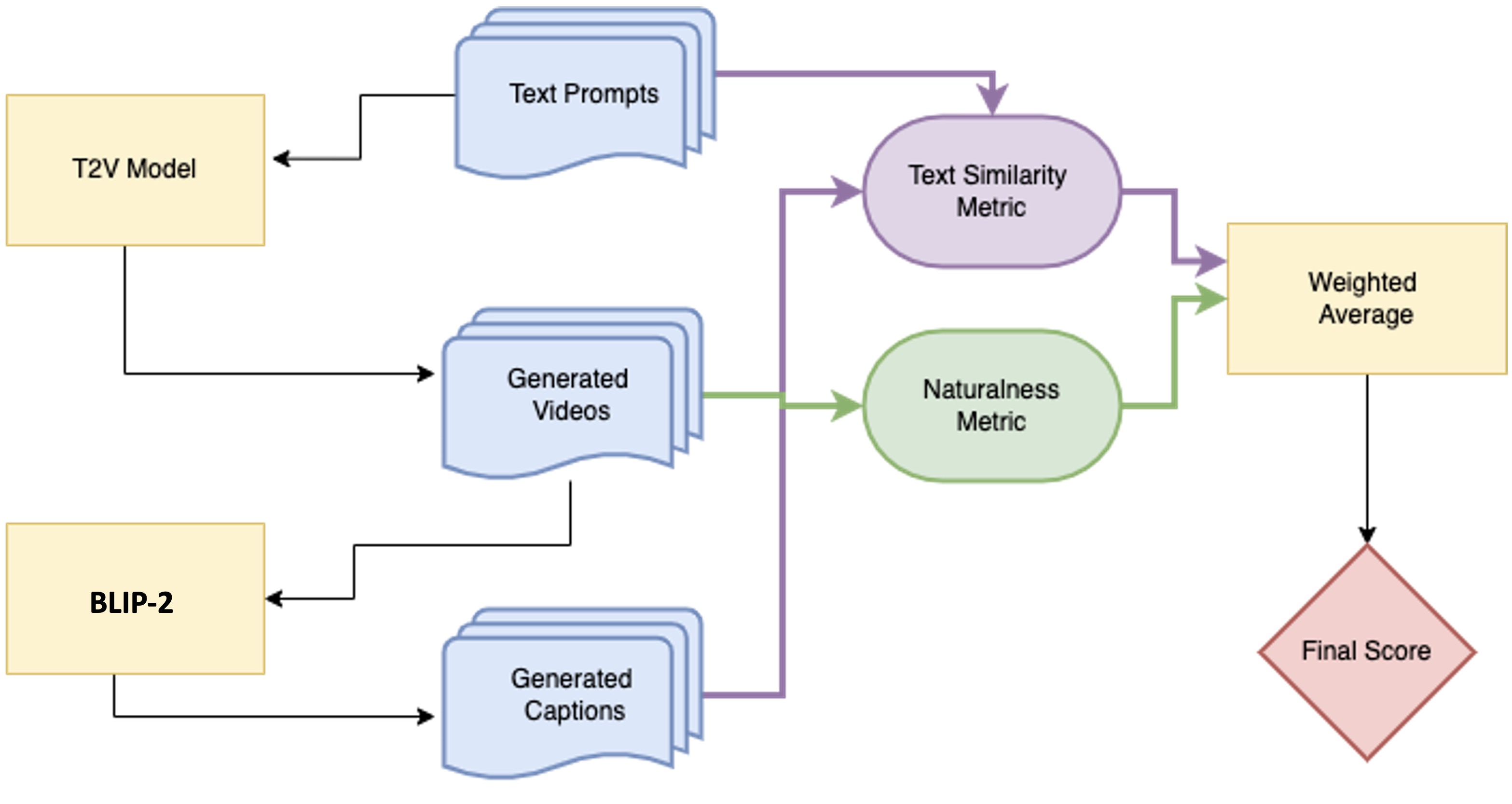}
    \caption{T2V-CL metric ensemble}
    \label{fig:metric_ensemble}
\end{figure}

To address the limitations of image naturalness and modal biases in  T2V bideo quality metrics we propose an ensemble with the workflow shown in Figure~\ref{fig:metric_ensemble}.
The first of the two parts involves data generation, depicted in blue and yellow boxes on the left side of the figure. Starting with an initial set of text prompts, we generate a video for each using a T2V model under evaluation.  The videos are used  to produce a set of captions for each using BLIP-2~\cite{li2023blip}.
The second part involves an ensemble of two metrics, the Text Similarity Metric which calculates a similarity score between the original text prompt and  BLIP-generated captions. Next, we use the Naturalness Metric, a customised XGBoost classifier that takes the generated video as input and outputs a score. Both metric outputs are in the range $[0,1]$.

\vspace{-10pt}

\noindent 
A weighted average based on a linear regression model trained using manually rated videos  described later is used to combine the  individual metrics. 

\subsection{Evaluating Image Naturalness}
\label{sec:Evaluating Image Naturalness}

We  developed a  classifier for video naturalness for which we collected and analysed several statistical measures from each video, including the following:
\begin{enumerate}
    \item \textbf{Texture score} measures the degree of uniformity in an image's texture since natural images, such as landscapes or animal fur, tend to have more complex textures than synthetic images.  After converting to grayscale and applying a Gaussian blur to reduce noise we apply Sobel edge detection in the x and y directions  and calculate the magnitude of the gradient. The variance of this magnitude is  the texture score. 
    
    \item \textbf{The sharpness score}  measures the amount of high-frequency content in an image, indicative of the image's level of detail. It is calculated by applying a sharpening filter to the image and then taking the RMS difference between the original  and the filtered image. 
    
    \item \textbf{The colour distribution score} is a measure of the uniformity of colour in an image, exploiting the characteristic of a uniform or artificial colour distribution in a non-natural image. It is calculated by applying K-means clustering with K=2 to the A and B channels of the image's LAB representation. This score is  the proportion of pixels in the cluster with the lowest A channel value.  
    
    \item \textbf{The spectral score} measures the extent to which an image differs from the natural image statistics in the Fourier domain. The function calculates the mean and standard deviation of each colour channel  and  computes the spectral score as the sum of standard deviations divided by the sum of means. 
    
    \item \textbf{The entropy score}  uses the Shannon entropy formula~\cite{cover2006elements}, which measures the level of randomness or disorder in pixel values.
   Natural images tend to have a higher degree of order and lower entropy than non-natural ones.
    
    \item \textbf{The contrast score} measures  differences between the lightest and darkest parts of an image by dividing the standard deviation of  pixel intensities by the mean intensity.
    
    \item \textbf{Oriented FAST and Rotated BRIEF (ORB)} is a feature detection algorithm~\cite{rublee2011orb} to compute statistics about the key points in an image including the mean and standard deviations of the distances between key points and  of the lengths of the descriptors associated with those key points. 
    
    \item \textbf{The number and sizes of blobs} is detected using the Laplacian of Gaussian (LoG) method~\cite{lindeberg1998feature}. Blobs are regions in an image with a relatively uniform intensity that stand out compared to the surrounding area.
 \end{enumerate}

Despite  poor performance on some images, NIQE and BRISQUE scores are useful in filtering  very noisy and disordered images thus they are included in our  metric comparisons. To facilitate processing, a YUV444 video frame is reshaped from planar  to interleaved format, which represents colour information in terms of brightness (Y) and colour (U and V), with 8 bits allocated to each channel. NIQE scores are calculated for the grayscale frame and for the Y, U and V channels in the YUV444 video frame separately as this provides a better visual representation of the image~\cite{podpora2014yuv} and  a more accurate evaluation of quality metrics like NIQE, which are more sensitive to variations in the chrominance channels. 

To train a classifier for image naturalness we also calculated a Modified Inception Score (MIS) for each video which operates on a similar principle to  Inception Score outlined in Section~\ref{sec:Evaluation Metrics} by calculating the mean probability distributions of all frames in a generated video. We modified the IS metric to return a larger value if the mean probability distribution in a video has low entropy.  Essentially, if the Inception model assigns a greater probability to one particular class throughout the frames in a video, MIS will produce a larger value. We achieved this by setting the marginal distribution to the uniform distribution. 

We collected all the video feature data described above from 187 videos comprising 92 natural and 95 non-natural. We approached the naturalness classifier task as a binary classification problem and manually assigned each video a label indicating natural or not. We trained three classifiers, AdaBoost, a Bagging classifier with a DecisionTree base and XGBoost. To optimise the performance of each classifier, we employed GridSearch. We evaluated the classifiers' performance using  F1  on  training, validation, and test sets. The XGBoost classifier performed the best on unseen data and was used in the rest of the paper.

\subsection{Evaluating Text Similarity}
\label{sec:nlp}

In the second part of our metric comparison we measure the  semantic similarity between the generated video captions and the original text prompt. The process involves generating captions for each video frame using BLIP-2 and measuring the similarity between each caption. 

In our approach we combine BERT and Cosine similarity. 
Using the example illustrated in \autoref{tab:sim-scores} we see that BERT tends to over-perform as is designed to capture more nuanced and complex semantic relationships between sentences or captions, whereas Cosine  only considers  surface-level similarity based on word overlap. By penalising the BERT similarity score with the Cosine similarity score, we ensure that the combined similarity  shown in \autoref{eq:combined_sim} reflects both  surface-level and deeper semantic similarities between two captions, thus providing a more accurate representation of their overall similarity. After conducting an analysis and running multiple experiments, we determined that the optimal ratio between BERT and Cosine similarities is 0.75~:~0.25.

\begin{table}[htb]
    \centering
    \begin{tabular}{l|ccc}
         &  ~~Cosine & BERT & Combined \\
         \midrule
    The sunrise was beautiful  over the ocean & \multirow{2}{3em}{~~0.00} & \multirow{2}{3em}{0.45} & \multirow{2}{3em}{0.22} \\
    The bulldozer was loud  and destroyed the building & & & \\
    \midrule
    A hot air balloon in the sky  & \multirow{2}{3em}{~~0.28} & \multirow{2}{3em}{0.76} & \multirow{2}{3em}{0.65} \\
    Balloon full of water exploding in extreme slow motion~~  & & & \\
    \bottomrule
    \end{tabular}
    \caption{Similarity scores for two pairs of sentences}
    \label{tab:sim-scores}
\end{table}

\vspace{-1.5cm}

\begin{equation}
    \text{Combined sim}=
    \begin{cases}
      0.25~( \text{Cos sim}) + 0.75~(\text{BERT sim}), & \ if ~\text{Cos sim} \neq 0 \\
      0.5~( \text{BERT sim}) & \ otherwise
    \end{cases}
\label{eq:combined_sim}
\end{equation}

\begin{figure}
    \centering
    \begin{subfigure}{0.15\textwidth}
    \centering
    \includegraphics[width=\textwidth]{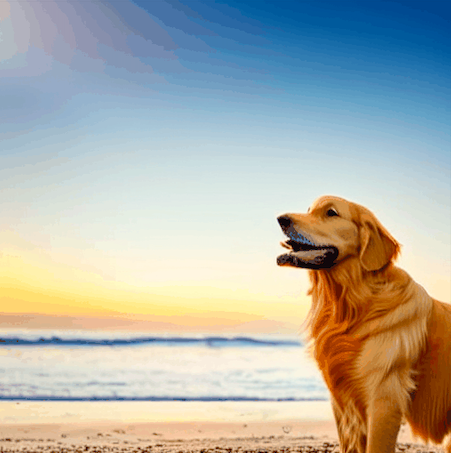}
    \end{subfigure}
    \begin{subfigure}{0.15\textwidth}
    \centering
    \includegraphics[width=\textwidth]{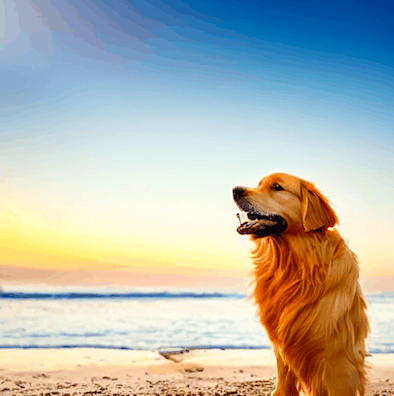}
    \end{subfigure}
    \begin{subfigure}{0.15\textwidth}
    \centering
    \includegraphics[width=\textwidth]{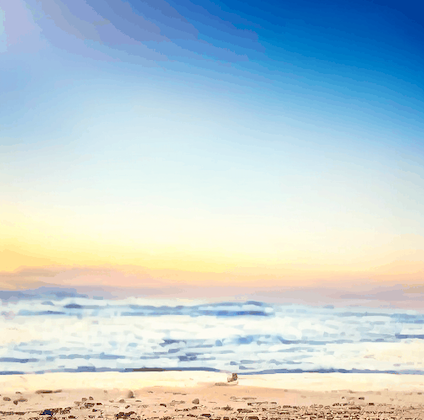}
    \end{subfigure}
    \begin{subfigure}{0.15\textwidth}
    \centering
    \includegraphics[width=\textwidth]{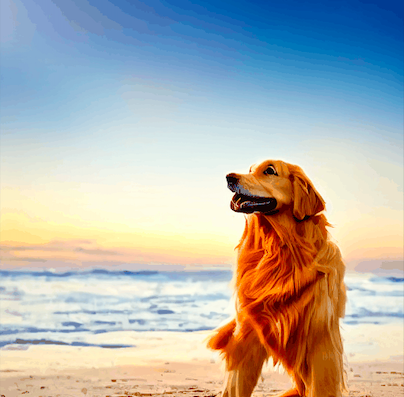}
    \end{subfigure}
    \begin{subfigure}{0.15\textwidth}
    \centering
    \includegraphics[width=\textwidth]{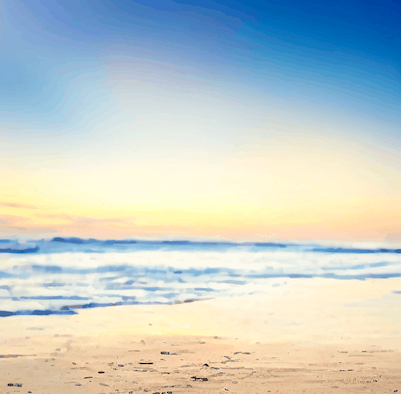}
    \end{subfigure}
    \begin{subfigure}{0.15\textwidth}
    \centering
    \includegraphics[width=\textwidth]{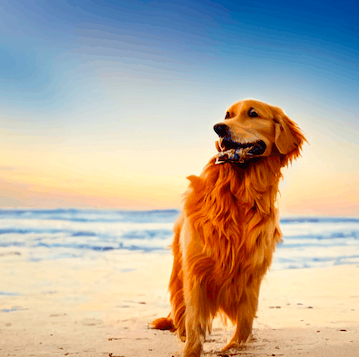}
    \end{subfigure}
    \centering
\caption{Frames from a generated video with the prompt ``A golden retriever eating ice cream on a beautiful tropical beach at sunset''. Note that 2 of the frames are missing the dog.}
\label{tab:TaV example}
\end{figure}

Given that some frames in  generated videos may exhibit significant distortions or omissions or not contain recognisable objects such as  in \autoref{tab:TaV example} where two frames do not include a dog, we calculate the weighted textual similarity for a generated video of $n$ frames as $\frac{1}{n}\sum_{i=1}^{n}w_{i}\cdot \text{sim}_{i}$. The weights are assigned based on the frequency of each caption in the overall list of generated captions. 

\section{Evaluation}
\label{sec:Evaluation}

\begin{figure}
    \centering
    \includegraphics[width=0.8\linewidth]{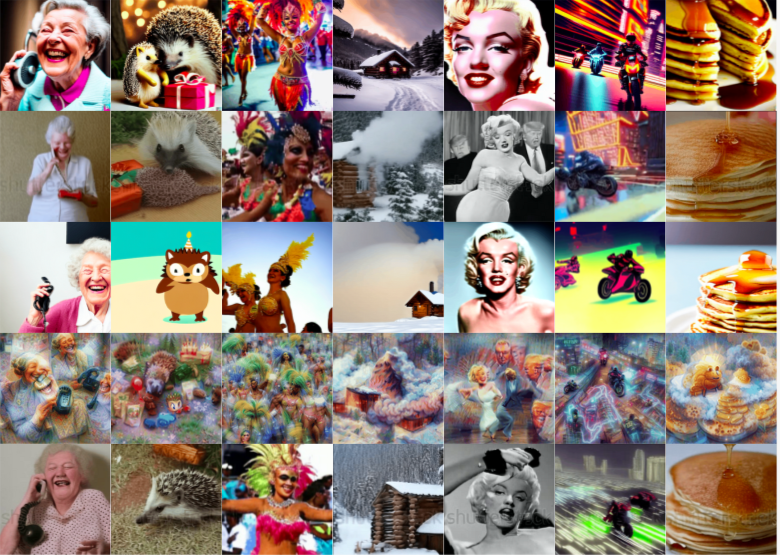}
    \caption{Samples from our generated videos -- rows show frames  generated by Text2Video-Zero, Text-to-Video Synthesis, Tune-a-Video, Aphantasia and Video Fusion respectively while  columns are  frames from the same text prompts. 
    \label{fig:collage}}
\end{figure}

We now present the comparison of quality metrics for videos generated from T2V models.
We used 201 prompts and  5 T2V models outlined earlier in Section~\ref{sec:Text-to-Video Models} to create 1,005 T2V model videos.
We carefully selected the 201 prompts by combining content generated by ChatGPT with manual curation. The compilation covers a broad range of topics including influential figures, notable places, and cultural events like Easter and the Brazilian Carnival. 87 of the prompts are short (4 to 8 words), 43 are of average length (9 to 13 words) and 71 are longer than 13 words. The  collection of prompts offers a diverse range of videos, spanning from practical scenarios to creative concepts. 
The videos encompass a variety of actions, relationships, and visual styles. 
Example frames from the collection of generated videos are shown in \autoref{fig:collage}. For each of the 1,005 generated videos we computed two measures, naturalness  and text similarity score as described earlier.

To obtain human quality scores for the generated videos we recruited volunteers to rate videos remotely and in their own time with each person given 10 days to complete the task across up to 10 sessions. The annotators rated each video on a scale of 1 (low) to 10 (high) for two aspects,  alignment and perception.
Alignment reflects the compatibility between the generated video and the text of the original prompt which was  provided, while perception rates the overall perceptual quality of the video considering issues such as clarity and sharpness, the presence of  visual artefacts, blurriness, or pixelation and whether colours represent real-world scenes and objects.  24 (16 male, 8 female) adult annotators, mostly graduate students, completed two ratings of each video giving  $1,005~ videos \times 24~ annotators \times 2~ aspects$ =
48,240 quality ratings.  Annotators were rewarded with a gift token when they  completed annotating the  videos.

In assessing the quality of still images the ``de facto'' metric  is  mean opinion score (MOS) \cite{streijl2016mean} which is the mean of the opinions and ratings of human evaluators gathered according to some numeric or qualitative scale. Despite its popularity  across various media including speech, audio, images, video, multimedia, etc. it does have  issues with its acceptability  \cite{streijl2016mean} because a single number cannot capture a diversity of opinion scores.
In \cite{6065690} the authors proposed  that the standard deviation of opinion scores reflects the subjective diversity while more recently \cite{3547872} proposed that as well as the mean of the opinion scores, researchers should assess quality in terms of the distribution of opinion scores, not just the standard deviation.

\begin{figure}[!htb]
    \centering
    \includegraphics[width=0.8\linewidth]{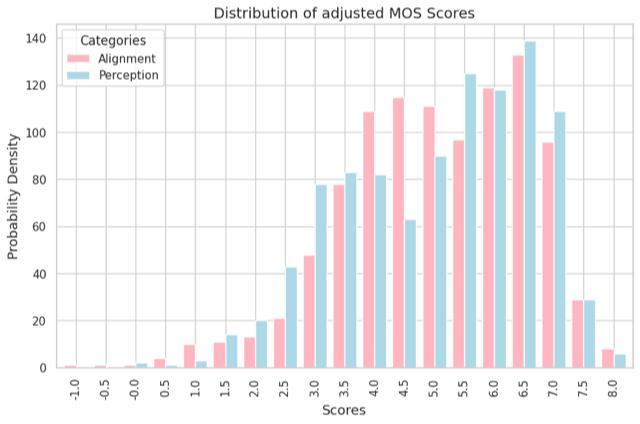}
    \caption{Distribution of adjusted MOS Scores.
    \label{fig:distribution_raw_mos_scores}}
\end{figure}

Following the approach taken by Li {\it et al.}~\cite{li2023agiqa} where the authors developed a quality metric for AI-generated text-to-image (T2I), we calculated a ``pid\_delta" by computing the difference between the average rating (5) and the mean of scores. We did this to see if the work in ~\cite{li2023agiqa} for T2I images could be applied to T2V videos and we refer to this as T2V-EC. Z-scores were computed by subtracting the mean from each score and dividing by the standard deviation and outliers were determined by comparing scores to a range defined by the mean and standard deviation. Re-scaling then restored the original data range by reversing the z-score transformation though negative values can result from the initial shift towards 5 during normalisation, especially where original scores were below 5.
Figure~\ref{fig:distribution_raw_mos_scores} shows the distribution of adjusted MOS scores for  alignment and perception. An overall human evaluation score for each video is the average adjusted MOS scores for alignment and perception. It is clear that  correlation between alignment and perception is not contingent upon the  rating values.

\begin{table}[!htb]
    \centering
\resizebox{\linewidth}{!}{
    \begin{tabular}{l|ccccc}
    \toprule
\multirow{2}{*}{Model}     &	\multicolumn{2}{c}{Alignment} & 	\multicolumn{2}{c}{Perception} & Combined\\				
	& ~~~Mean~~~	& ~~~Std.Dev~~~	 & ~~~Mean~~~	& ~~~Std.Dev~~~		 &	Human Score \\
\midrule
Aphantasia &	4.016&	0.841	&3.221&	0.692	&	0.362 \\	
Text2Video-Zero~~~~	&5.985	&1.139	&6.393	&0.886		&0.619 \\
T2VSynthesis&	5.333	&1.622	&5.485	&1.366	&0.541 \\
Tune-a-Video&	5.053	&1.340	&5.070	&1.196	&0.506 \\		
Video\_Fusion&	4.995	&1.686	&5.139	&1.507	&0.507 \\
\bottomrule
    \end{tabular}
    }
    \caption{Human evaluation scores for five T2V models.
    \label{tab:MOS-stats}}
\end{table}

\begin{figure}[!htb]
    \centering
\includegraphics[width=\linewidth]{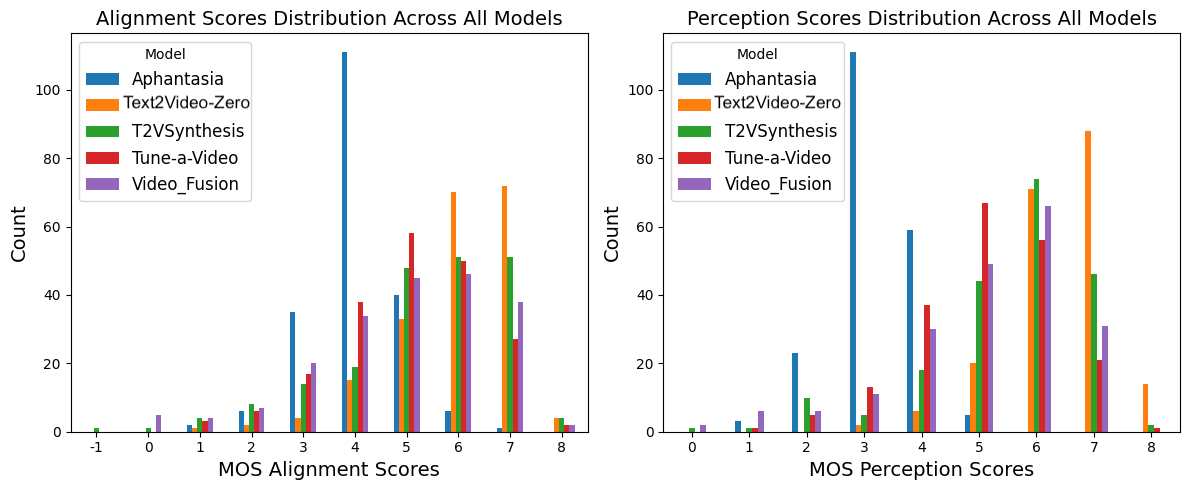}
    \caption{MOS score distributions across different models.
    \label{fig:models_dist}}
\end{figure}

\begin{figure}[!htb]
    \centering
    \includegraphics[width=0.9\linewidth]{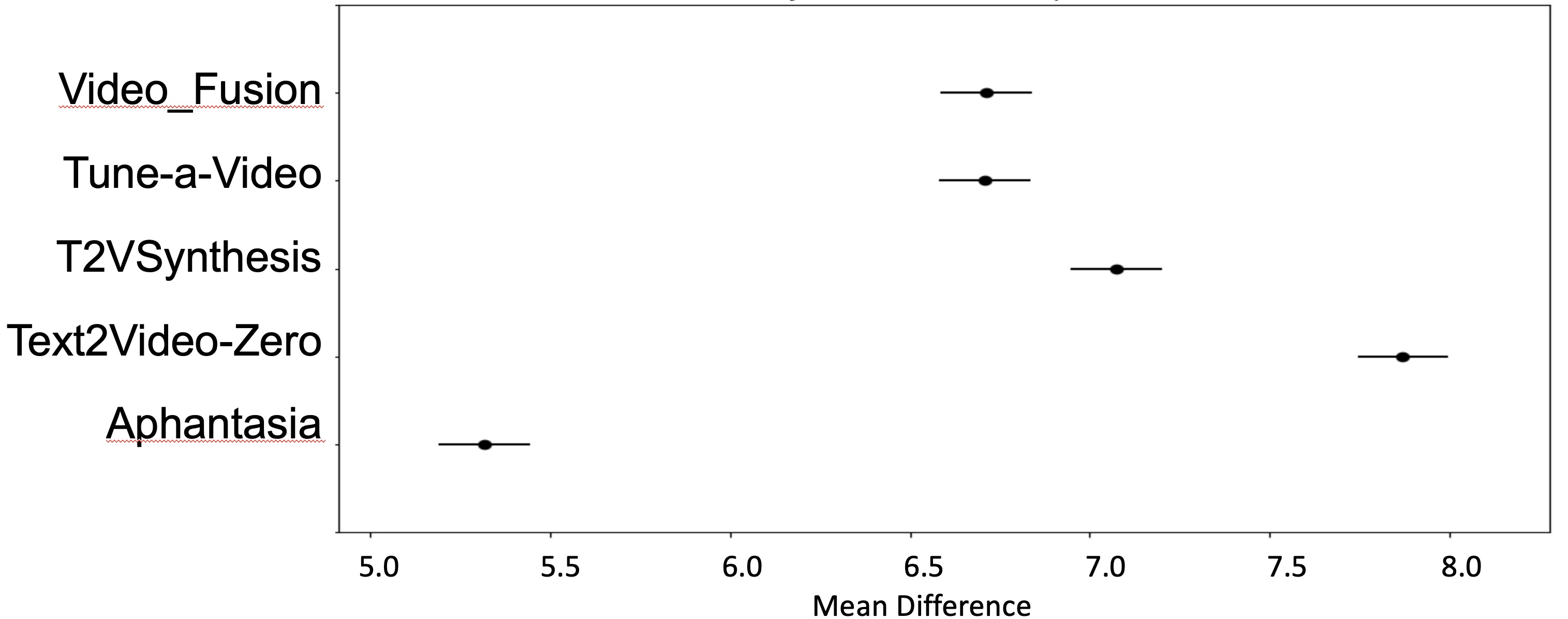}
    \caption{Tukey HSD of adjusted MOS Scores  across different models.
    \label{fig:turkey_hsd}}
\end{figure}

We now examine how  human evaluation differentiated among the 5 models for alignment, perception  and for the combined human score (average), and these are shown as  means and standard deviations in Table~\ref{tab:MOS-stats} and as distributions in Figure~\ref{fig:models_dist}.
What these  show is that Aphantasia is the worst-performing across perception and alignment, Text2Video-Zero is best and Tune-a-Video, Video Fusion, and Text2Video Synthesis appear to be about the same
To support this interpretation we conducted a Tukey HSD (honestly significant difference) test on the adjusted combined human evaluation scores to identify differences among the 5 models. 
Tukey's HSD \cite{abdi2010tukey} is a significance test to simultaneously compare multiple means and distributions, MOS scores for 5 T2V models in our case, in a single step and to find means whose  differences are greater than the expected standard error.
The results of the test are shown in  \autoref{fig:turkey_hsd} and show that Aphantasia exhibits notably lower performance than the others while Text2Video-Zero demonstrates significantly higher performance.  The other three models --- Tune-a-Video, Video\_Fusion, and Text2Video Synthesis —-- exhibit relatively similar levels of performance.

\begin{figure}[!htb]
    \centering
    \includegraphics[width=0.8\linewidth]{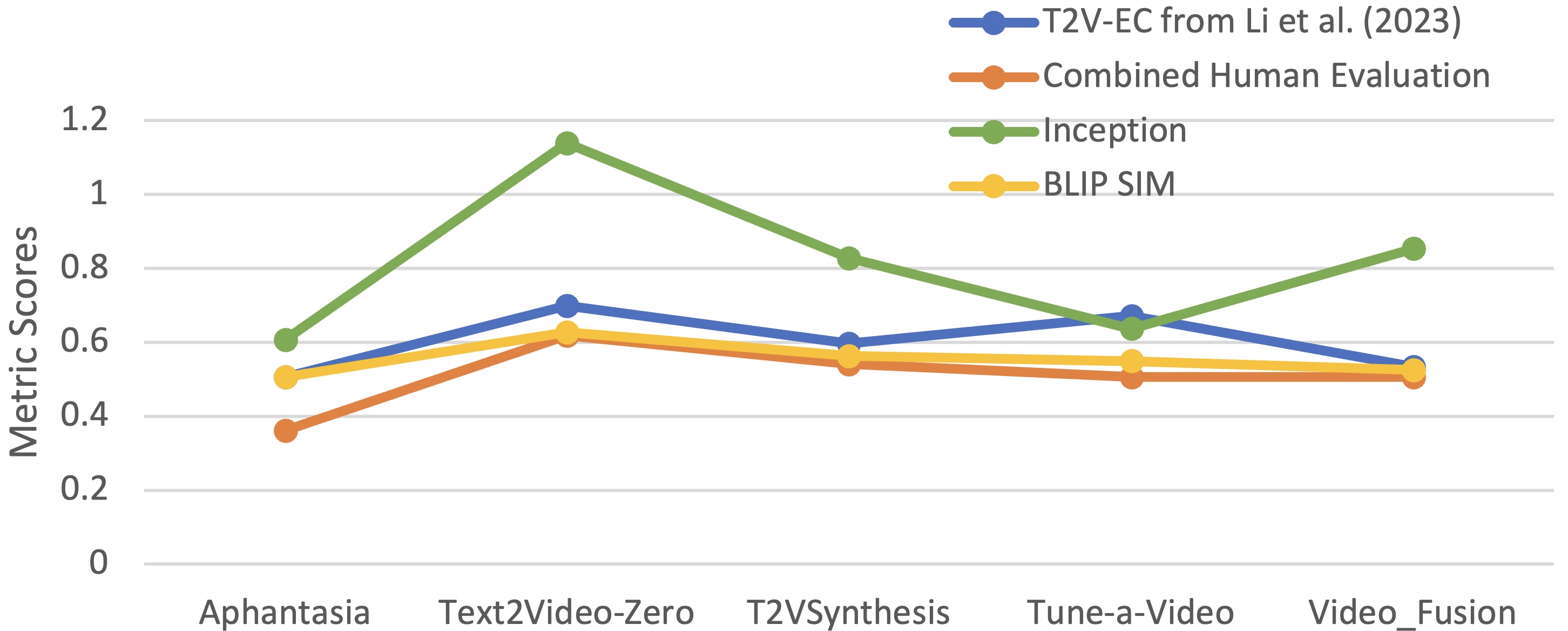}
    \caption{Comparison of model ranks by different metrics.
    \label{fig:metric_figs}}
\end{figure}

The combined human evaluation scores were compared against  T2V-CL replicating the work of Li {\it et al.}~\cite{li2023agiqa} for images, BLIPSim and Inception scores  in Figure~\ref{fig:metric_figs}.
The replicated T2V-CL metric shows a consistency in ranking with the other metrics except for the Tune-a-Video model which has a higher rank by compared to human evaluation. This discrepancy is because Tune-a-Video predominantly produces cartoon-style videos, as shown in \autoref{fig:collage} and our naturalness classifier was trained on a range of example videos from a range of models which did not include enough from the Tune-a-Video model.

\begin{figure}[!htb]
    \centering
    \includegraphics[width=0.9\linewidth]{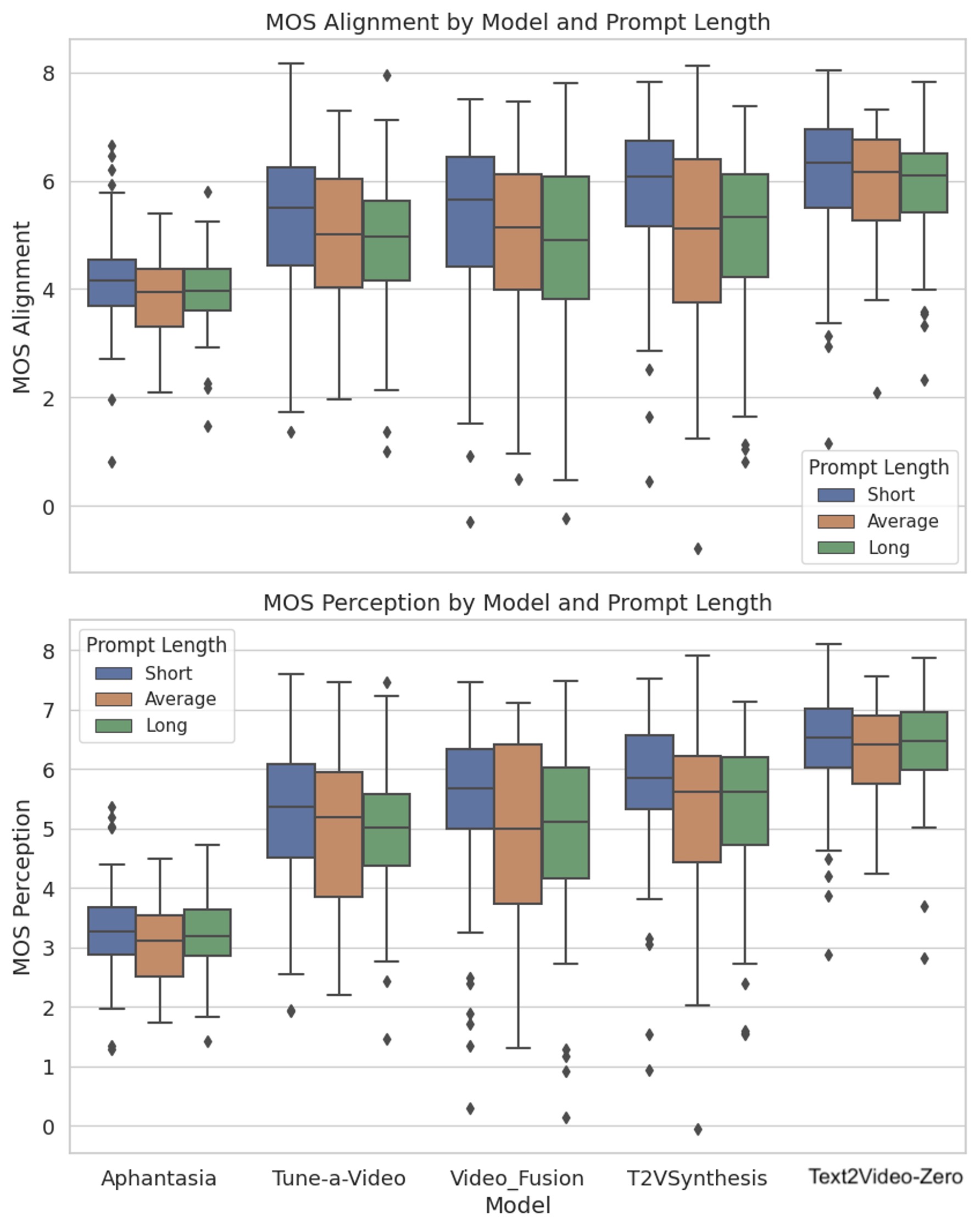}
    \caption{Boxplots of adjusted alignment (top) and perception (bottom) scores across different prompt lengths
    \label{fig:combined-box}}
\end{figure}

In a final analysis we examined how prompt length influences model performance. Figure~\ref{fig:combined-box} shows boxplots of adjusted MOS scores indicating the ordering of model performance with Text2Video-Zero best and Aphantasia worst in both boxplots and for all prompt lengths. They also show that almost always the shorter the prompt length, the better the quality of the video. This is explained by the fact that alignment to a longer prompt is more difficult for a T2V model.

\section{Conclusions}
\label{sec:Conclusion}

We investigated techniques for evaluating the quality of text-to-video (T2V) model outputs including a critical analysis of commonly-employed  metrics. We examined the limitations of existing methods for evaluating video quality with an emphasis on assessing the naturalness of T2V  content as well as the degree of semantic correspondence between  videos and  text  prompts used to generate them. We provide  an open dataset of T2V videos from 5 models, with human annotations of their quality\footnote{Code, prompts, examples and the 1,005 generated videos from the 5 models used in this paper are  available for public access at \url{https://tinyurl.com/4eufrek8}}. 
In summary we can say that there is some consistency across the metrics and with human evaluations but not yet at the level where we can eliminate the need for expensive and time-consuming human assessment, though  automatic metrics are a good proxy.

%
%
%


\bibliographystyle{splncs04}
\bibliography{bibliography}

%
%
%
%
\end{document}